\documentclass[letterpaper]{article} 
\usepackage{aaai24}  
\usepackage{times}  
\usepackage{helvet}  
\usepackage{courier}  
\usepackage[hyphens]{url}  
\usepackage{graphicx} 
\usepackage{natbib}  
\usepackage{caption} 
\usepackage{algorithm}
\usepackage{algorithmic}

\usepackage{subfig}
\usepackage{amsfonts,amssymb}
\usepackage{multirow}
\usepackage{booktabs}
\usepackage{pifont}
\usepackage{color}
\usepackage{array}
\usepackage{listings}
\newcommand{\vpara}[1]{\noindent\textbf{#1}}

\usepackage{newfloat}
\usepackage{listings}
\DeclareCaptionStyle{ruled}{labelfont=normalfont,labelsep=colon,strut=off} 
\floatstyle{ruled}
\newfloat{listing}{tb}{lst}{}
\floatname{listing}{Listing}
\title{Layer-wise Representation Fusion for Compositional Generalization}
\author{
    Yafang Zheng\textsuperscript{\rm 1,2}\equalcontrib, Lei Lin\textsuperscript{\rm 1,2,3}\equalcontrib, Shuangtao Li\textsuperscript{\rm 1,2}, Yuxuan Yuan\textsuperscript{\rm 1,2}, Zhaohong Lai\textsuperscript{\rm 1,2}, \\ Shan Liu\textsuperscript{\rm 1,2}, Biao Fu\textsuperscript{\rm 1,2}, Yidong Chen\textsuperscript{\rm 1,2}, Xiaodong Shi\textsuperscript{\rm 1,2}\thanks{Corresponding Author.}
}
\affiliations{
    \textsuperscript{\rm 1}Department of Artificial Intelligence, School of Informatics, Xiamen University\\
    \textsuperscript{\rm 2} Key Laboratory of Digital Protection and Intelligent Processing of Intangible Cultural Heritage\\of Fujian and Taiwan (Xiamen University), Ministry of Culture and Tourism, China\\
    \textsuperscript{\rm 3}Kuaishou Technology, Beijing, China\\
    \{zhengyafang, linlei\}@stu.xmu.edu.cn, \{ydchen, mandel\}@xmu.edu.cn \\

}

\usepackage{bibentry}

\begin{document}

\maketitle

\begin{abstract}
Existing neural models are demonstrated to struggle with compositional generalization (CG), i.e., the ability to systematically generalize to unseen compositions of seen components. A key reason for failure on CG is that the syntactic and semantic representations of sequences in both the uppermost layer of the encoder and decoder are entangled. However, previous work concentrates on separating the learning of syntax and semantics instead of exploring the reasons behind the representation entanglement (RE) problem to solve it.
We explain why it exists by analyzing the representation evolving mechanism from the bottom to the top of the Transformer layers. We find that the ``shallow'' residual connections within each layer fail to fuse previous layers' information effectively, leading to information forgetting between layers and further the RE problems. Inspired by this, we propose LRF, a novel \textbf{L}ayer-wise \textbf{R}epresentation \textbf{F}usion framework for CG, which learns to fuse previous layers' information back into the encoding and decoding process effectively through introducing a \emph{fuse-attention module} at each encoder and decoder layer. LRF achieves promising results on two realistic benchmarks, empirically demonstrating the effectiveness of our proposal. Codes are available at~\url{https://github.com/thinkaboutzero/LRF}.
\end{abstract}

\section{Introduction}
\label{sec:intro}
The remarkable progress of sequence-to-sequence (seq2seq) models in language modeling has been primarily attributed to their ability to learn intricate patterns and representations from vast amounts of data~\cite{sutskever2014sequence,dong2016language,vaswani2017attention}.
However, a critical challenge that remains unsolved for neural sequence models is the ability to understand and produce novel combinations from known components~\cite{fodor1988connectionism,lake2017building}, i.e., \textit{compositional generalization} (CG). For example, if a person knows ``the doctor has lunch'' [Der Arzt hat Mittagessen] and ``the lawyer'' [Der Anwalt] where the segment in ``[]'' denotes the German translation, then it is natural for the person to know the translation of ``the lawyer has lunch'' is [Der Anwalt hat Mittagessen] even though they have never seen it before. Such nature is beneficial for models to perform robustly in real-world scenarios, as even huge training data can not cover a potentially infinite number of novel combinations.

\begin{figure}[!t]
    \centering
    \subfloat[Standard Transformer\label{fig:fig1-1}]{
    
    \includegraphics[width=0.4\linewidth]{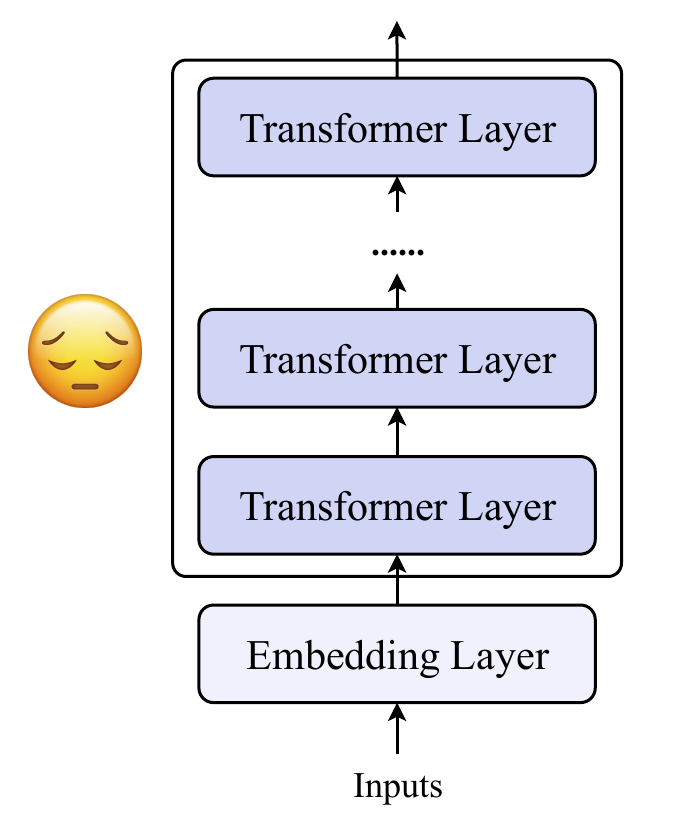}
    }
    \subfloat[LRF (Ours)\label{fig:fig1-2}]{
    \includegraphics[width=0.4\linewidth]{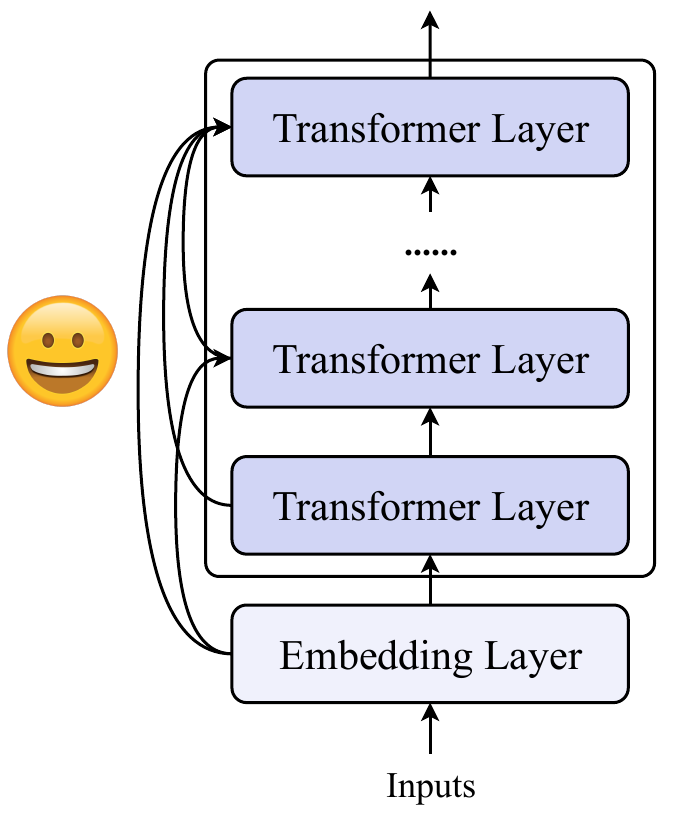}
    }
    \caption{Comparison between (a) Standard Transformer and (b) LRF (ours). (a) illustrates the standard architecture of the Transformer model and (b) illustrates the architecture of our method, which fuses representations from previous layers at each layer effectively.}
    \label{fig:fig1}
\end{figure}

Recent studies have demonstrated that a key reason for failure on CG is that the syntactic and semantic representations of sequences in both the uppermost layer of the encoder and decoder are entangled (i.e., encoder or decoder RE problem)~\cite{russin2019compositional,li2019compositional,raunak2019compositionality,liu2020compositional,DBLP:conf/acl/LiuALLCLWZZ21,DBLP:conf/iclr/MittalRRBL22,zheng2021disentangled,DBLP:conf/emnlp/ArthurNN16,thrush2020compositional,akyurek2021lexicon,DBLP:conf/emnlp/YaoK22,yin2022categorizing}. To alleviate it, one line of research on CG concentrates on utilizing separate syntactic and semantic representations. Specifically, they either produce two separate syntactic and semantic representations and then compose them appropriately~\cite{li2019compositional,russin2019compositional,JiangB21indu}, or design external modules and then employ a multi-stage generation process~\cite{liuqian2020compositional,DBLP:conf/acl/LiuALLCLWZZ21,corrabs220210745,corrabs221001603,eaclCazzaroLQC23}. 

However, they focus on separating the learning of syntax and semantics. Different from them, we explore the reasons behind the RE problem to solve it. We get insights from the findings that the residual connections are ``shallow'', and conclusions on analyzing the Transformer~\cite{peters-etal-2018-deep,he2019hard,voita2019bottom,belinkov2020linguistic} show that the bottom layers of the Transformer contain more syntactic information while the top ones contain more semantic information.
As shown in Figure~\ref{fig:fig1-1}, the residual connections within each layer have been ``shallow'' themselves, and only pass through simple, one-step operations~\cite{yu2018deep}, which make the model ``forget'' distant layers and fail to fuse information from previous layers effectively~\cite{bapna2018training,dou2018exploiting,wang2020multi,wang2019learning}. 
Furthermore, it is clear that the nature of the RE problems are lacking effective fusion of syntactic and semantic information stored in different layers, since the representations are gradually entangled from bottom to top of the Transformer layers~\cite{raunak2019compositionality,russin2019compositional,zheng2021disentangled}. 

Based on the above findings, we hypothesize that ``shallow'' residual connections within each layer are one of the reasons resulting in the RE problems.
To this end, we propose LRF, a novel \textbf{L}ayer-wise \textbf{R}epresentation \textbf{F}usion framework for CG, which learns to fuse previous layers' information at each layer effectively. 
Specifically, we extend the base model by fusing previous layers' information back into the encoding and decoding process by introducing a fuse-attention module in each encoder and decoder layer.

Experimental results on CFQ~\cite{keysers2019measuring} (semantic parsing) and CoGnition~\cite{li2021compositional} (machine translation, MT) empirically show that our method achieves better generalization performance, outperforming competitive baselines and other techniques. Notably, LRF achieves \textbf{20.0\%} and \textbf{50.3\%} (about \textbf{30\%}, \textbf{20\%} relative improvements) for instance-level and aggregate-level error reduction rates on CoGnition. Extensive analyses demonstrate that fusing previous layers' information at each layer effectively leads to better generalization results, outperforming competitive baselines and more specialized techniques.

\section{Methodology}
\label{sec:method}
We adopt the Transformer~\cite{vaswani2017attention} as an example for clarification, however, \textbf{our proposed method is applicable to any seq2seq models}. In the following, we first introduce the Transformer, and then our proposed LRF.

\subsection{Transformer}
\label{sec:transformer}
Given a sequence of a source sentence and a target sentence $X=\{x_{1}, ..., x_{S}\}, Y=\{y_{1}, ..., y_{T}\}$,
where $S,T$ denote the number of source and target tokens, respectively. The Transformer encoder first maps $X$ to embedding matrix $H^{0}$, and then takes $H^{0}$ as input and outputs a contextualized representation $H^{L} \in \mathbb{R}^{d \times S}$, where $d, L$ denote the hidden size and the number of encoder layers respectively. Similarly, the Transformer decoder maps $Y$ to embedding matrix $\hat{H}^{0}$ first, and then takes $\hat{H}^{0}$ as input and outputs a contextualized representation $\hat{H}^{L} \in \mathbb{R}^{d \times T}$.

\begin{figure}[!t]
    \centering
    \includegraphics[width=1.0\linewidth]{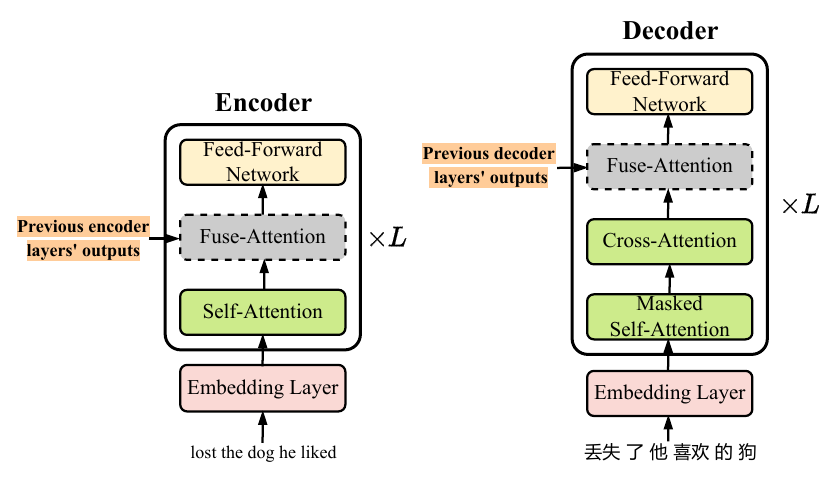}
    \caption{Architecture of LRF based on the Transformer. The dotted boxes denote the fuse-attention module.}
    \label{fig:fig2}
\end{figure}
\vpara{Attention Mechanism.}
An attention function can be described as a query ($Q$) and a set of key-value ($K$-$V$) pairs mapped to an output. Formally, given $Q$, $K$, and $V$, the scaled dot product attention mechanism is computed as:
\begin{equation}
\small
    {\rm Attention}(Q, K, V) = {\rm softmax}(\frac{Q^\top K}{\sqrt{d_{k}}})V,
\end{equation}
where $d_{k}$ is the dimension of $K$. 

A typical extension of the above is multi-head attention (MHA), where multiple linear projections are executed in parallel. The calculation process is as follows:
\begin{equation}
\small
    {\rm MHA}(Q, K, V) = [head_{1}; ...;head_{h}] W^{O},
\end{equation}
\begin{equation}
\small
    head_{i} = {\rm Attention}(QW_{i}^{Q}, KW_{i}^{K}, VW_{i}^{V}),
\end{equation}
where $W_{i}^{Q} \in \mathbb{R}^{d \times d_{k}}$, $W_{i}^{K} \in \mathbb{R}^{d \times d_{k}}$, $W_{i}^{V} \in \mathbb{R}^{d \times d_{v}}$ and $W_{i}^{O} \in \mathbb{R}^{hd_{v} \times d}$ are model parameters. $h$ denotes the number of heads.

\vpara{Layer Structure.} The Transformer encoder has $L$ identical layers, and each layer consists of two sub-layers (i.e., self-attention and feed-forward networks). The Transformer decoder has $L$ identical layers, and each layer consists of three sub-layers (i.e., masked self-attention, cross-attention and feed-forward networks). In the $l$-th self-attention layer of the encoder, the query, key and value are all the hidden states outputted by the previous layer $H^{l-1}$. The self-attention mechanism in the decoder operates in a similar manner. The formal expression is as follows:
\begin{equation}
\small
    H_{a}^{l} = {\rm MHA}(H^{l-1}, H^{l-1}, H^{l-1}),
\end{equation}
\begin{equation}
\small
    \hat{H}_{a}^{l} = {\rm MHA}(\hat{H}^{l-1}, \hat{H}^{l-1}, \hat{H}^{l-1}).
\end{equation}

In the $l$-th cross-attention layer of the decoder, the query is the hidden states outputted by the $l$-th self-attention layer $\hat{H}_{a}^{l}$ and the key and value are all hidden states outputted by the uppermost layer of the encoder $H^{L}$. The computation process is as follows:
\begin{equation}
\small
    \hat{H}_{ca}^{l} = {\rm MHA}(\hat{H}_{a}^{l}, H^{L}, H^{L}).
\end{equation}

The feed-forward sub-layer is a two-layer transformation with a ReLU activation function:
\begin{equation}
\small
    \label{eq7}
    H_{l} = W_{2}^{l} {\rm ReLU}(W_{1}^{l}H_{a}^{l} + b_{1}^{l}) + b_{2}^{l},
\end{equation}
\begin{equation}
\small
    \label{eq8}
    \hat{H}_{l} = W_{4}^{l} {\rm ReLU}(W_{3}^{l}\hat{H}_{ca}^{l} + b_{3}^{l}) + b_{4}^{l},
\end{equation}
where $W_{1}^{l}, b_{1}^{l}, W_{2}^{l}, W_{3}^{l}, b_{2}^{l}, W_{4}^{l}, b_{3}^{l}$ and $b_{4}^{l}$ are all trainable model parameters. We omit the layer normalization and residual connection for brevity.

\subsection{Layer-wise Representation Fusion (LRF)}
\label{sec:fusion}

Our proposed LRF extends the Transformer by introducing a fuse-attention module under the feed-forward module in each encoder and decoder layer, which learns to fuse information from previous layers at each encoder and decoder layer effectively.

\vpara{Fuse-attention Module.}
\label{sec:ia}
In the $l$-th encoder layer, all previous layers' outputs are stacked as $H_{plo} \in \mathbb{R}^{d \times l}$, where $l$ is the number of previous layers (include the embedding layer). The fuse-attention module fuses different aspects of language information from previous layers effectively via the multi-head attention mechanism:
\begin{equation}
\small
    H_{p}^{l} = {\rm MHA}(H_{a}^{l}, H_{plo}, H_{plo}),
\end{equation}
where $H_{plo} = \{H_{0}, ..., H_{l-1} \}$, and the localness of fuse-attention module is implemented by mask mechanisms. The output $H_{p}^{l}$ is fed into the $l$-th feed-forward sub-layer of the encoder (Eq.~\ref{eq7}).

In the $l$-th decoder layer, all previous layers' outputs are stacked as $\hat{H}_{plo} \in \mathbb{R}^{d \times l}$, where $l$ is the number of previous layers (include the embedding layer). The fuse-attention module fuses different aspects of language information from previous layers effectively via the multi-head attention mechanism:
\begin{equation}
\small
    \hat{H}_{p}^{l} = {\rm MHA}(\hat{H}_{ca}^{l}, \hat{H}_{plo}, \hat{H}_{plo}),
\end{equation}
where $\hat{H}_{plo} = \{\hat{H}_{0}, ..., \hat{H}_{l-1} \}$, and the localness of fuse-attention module is implemented by mask mechanisms. The output $\hat{H}_{p}^{l}$ is fed into the $l$-th feed-forward sub-layer of the decoder (Eq.~\ref{eq8}).

The differences between LRF and Transformer are illustrated by the dotted boxes in Figure~\ref{fig:fig2}. By introducing a fuse-attention module in every encoder and decoder layer, each layer of the encoder and decoder are able to access and fuse previous layers' information effectively.

\vpara{Training.}
\label{sec:ta}
Formally, $\mathcal{D} = \{(X, Y)\}$ denotes the training corpus, $\mathcal{V}$ denotes the vocabulary of $\mathcal{D}$. LRF aims to estimate the conditional probability $p(y_{1}, ..., y_{T} | x_{1}, ..., x_{S})$, where $(x_{1}, ..., x_{S})$ is an input sequence and $(y_{1}, ..., y_{T})$ is its corresponding output sequence:
\begin{equation}
\small
    p(Y | X; \{\theta^{(0)} \cup \theta_{+}\}) = \prod_{t=1}^{T+1} p(y_{t} | y_{< t}, X; \{\theta^{(0)} \cup \theta_{+}\}),
\end{equation}
where $t$ is the index of each time step, $y_{< t}$ denotes a prefix of $Y$,  $\theta^{(0)}$ denotes initial parameters of the Transformer model, $\theta_{+}$ denotes parameters of the fuse-attention modules and each factor $p(y_{t} | X, y_{1}, ..., y_{t-1};$ $\{\theta^{(0)} \cup \theta_{+}\})$ is defined as a ${\rm softmax}$ distribution of $\mathcal{V}$.

During training, the model is optimized using cross-entropy (CE) loss, which is calculated as follows:
\begin{equation}
\small
    L_{CE}(\{\theta^{(0)} \cup \theta_{+}\}) = -\sum_{t=1}^{T+1}\log p(y_{t} | y_{< t}, X; \{\theta^{(0)} \cup \theta_{+}\}).
\end{equation}

\section{Experiments}
\label{sec:ex}

We mainly evaluate LRF on two comprehensive and realistic benchmarks for measuring CG, including semantic parsing (\textit{CFQ}) and machine translation (\textit{CoGnition}).

\subsection{Experimental Settings}
\label{sec:est}

\begin{table*}[t]
\centering
    \small
\resizebox{1.0\linewidth}{!}{
\begin{tabular}{cccccccc}
\toprule
\multirow{2}{*}{\bf Model} & \multirow{2}{*}{\bf Params} & \multicolumn{5}{c}{\bf Compound Translation Error Rate (CTER) $\downarrow$} & \multirow{2}{*}{\bf BLEU $\uparrow$}\\
\cmidrule{3-7}
& & NP & VP & PP & Total & $\Delta$ \\
\midrule
Transformer & 35M & 24.7\%/55.2\% & 24.8\%/59.5\% & 35.7\%/73.9\% & 28.4\%/62.9\% & -/- & 59.5 \\

Transformer-Rela & 35M & 30.1\%/58.1\% & 27.6\%/61.2\% & 38.5\%/74.1\% & 32.1\%/64.5\% & +3.7\%/+1.6\% & 59.1 \\

Transformer-Small & 25M & 25.1\%/56.9\%  & 25.6\%/60.3\% & 39.1\%/75.0\% & 29.9\%/64.5\% & +1.5\%/+1.6\% & 59.0 \\

Transformer-Deep & 46M & 21.4\%/51.4\%  & 23.2\%/57.6\% & 32.5\%/71.5\% & 25.7\%/60.2\% & -2.7\%/-2.7\% & 60.2 \\

\midrule
Bow & 35M & 22.2\%47.9\% & 24.8\%/55.6\% & 	35.0\%/73.2\% & 27.3\%/58.9\% & -1.1\%/-3.0\% & - \\

SeqMix & 35M & 24.5\%/49.7\% & 26.9\%/58.9\% & 34.4\%/73.1\% & 28.6\%/60.6\% & +0.2\%/-2.3\% & - \\

Dangle & 35M & -/-  & -/- & -/- & 24.4\%/55.5\% & -5.0\%/-7.4\% & 59.7 \\

Proto-Transformer & ~42M & 14.1\%/36.5\%  & 22.1\%/50.9\% & 28.9\%/68.2\% & 21.7\%/51.8\% & -6.7\%/-11.1\% & 60.1 \\
Transformer+CReg & 25M & -/- & -/- & -/- & 20.2\%/\textbf{48.3\%} & -8.2\%/\textbf{-14.6\%} & 61.3 \\
DLCL & 35M & -/- & -/- & -/- & 28.4\%/67.9\% & +0.0\%/+5.0\% & 59.2 \\
\midrule
LRF & 48M & \bf 12.0\%/34.5\%  & 20.9\%/50.7\% & 27.2\%/65.8\% & \textbf{20.0\%}/50.3\% & \textbf{-8.4\%}/-12.6\% & 61.8 \\

LRF-Rela & 48M & 12.9\%/36.2\%  & \bf 20.5\%/50.6\% & \bf 26.9\%/65.0\% & 20.1\%/50.6\% & -8.3\%/-12.3\% & \bf 62.0 \\

LRF-Small & 33M & 14.1\%/38.8\%  & 21.4\%/51.7\% & 27.9\%/65.6\% & 21.1\%/52.1\% & -7.3\%/-10.8\% & 60.9 \\
\bottomrule
\end{tabular}
}
\caption{
CTERs (\%) on CoGnition. We report instance-level and aggregate-level CTERs in the CG-test set, separated by ``/''. In addition, we also report the commonly used metric BLEU score in MT tasks. ``-'' denotes that the results are not provided in the original paper. Results are averaged over 6 random runs.
}
\label{table:t2}
\end{table*}

\begin{figure}[!t]
    \centering
    \includegraphics[width=1.0\linewidth]{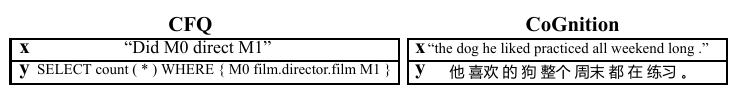}
    \caption{Examples of CFQ and CoGnition.}
    \label{fig:fig3}
\end{figure}
\vpara{Datasets.} \textit{CoGnition} is an English $\rightarrow$ Chinese (En$\rightarrow$Zh) translation dataset, which is used to systematically evaluate CG in MT scenarios. It consists of a training set of 196,246 sentence pairs, a validation set and a test set of 10,000 samples. In particular, it also has a dedicated synthetic test set (i.e., CG-test set) consisting of 10,800 sentences containing novel compounds, so that the ratio of compounds that are correctly translated can be computed to evaluate the model’s ability of CG directly. 
\textit{CFQ} is automatically generated from a set of rules in a way that precisely tracks which rules (atoms) and rule combinations (compounds) of each example. In this way, we can generate three splits with \emph{maximum compound divergence} (MCD) while guaranteeing a small atom divergence between train and test sets, where large compound divergence denotes the test set involves more examples with unseen syntactic structures. We evaluate our method on all three splits. Each split dataset consists of a training set of 95,743, a validation set and a test set of 11,968 examples. Figure~\ref{fig:fig3} shows examples of them. 

\vpara{Data.} We follow the same settings of~\cite{li2021compositional} and~\cite{keysers2019measuring} to preprocess CoGnition and CFQ datasets separately. For CoGnition, we use an open-source Chinese tokenizer\footnote{https://github.com/fxsjy/jieba} to preprocess Chinese and apply Moses tokenizer\footnote{https://github.com/moses-smt/ \\ mosesdecoder/blob/master/scripts/tokenizer/tokenizer.perl} to preprocess English, which is the same in~\cite{lin23leapt}. 
For CFQ, we use the GPT2-BPE tokenizer\footnote{https://github.com/facebookresearch/fairseq/blob/main/examples/ \\ roberta/multiprocessing\_bpe\_encoder.py} to preprocess source and target English text.

\vpara{Setup.} For CoGnition, we follow the same experimental settings and configurations of~\cite{li2021compositional}. We build our model on top of Transformer~\cite{vaswani2017attention}. 
We use one GeForce GTX 2080Ti for training with 100,000 steps and decoding. For CFQ, we follow the same experimental settings and configurations of~\cite{zheng2021disentangled}. We build our model on top of RoBERTa~\cite{liu2019roberta}. We use the base RoBERTa with 12 encoder layers, which is combined with a Transformer decoder that has 2 decoder layers with hidden size 256 and feed-forward dimension 512. We use one GeForce GTX 2080Ti for training with 45,000 steps and decoding. We implement all comparison models and LRF with an open source Fairseq toolkit~\cite{ott2019fairseq}. 

\vpara{Evaluation Metrics.} For CoGnition, we use compound translation error rate (CTER~\cite{li2021compositional}) to measure the model’s ability of CG. Specifically, \textit{instance-level} CTER (i.e.,$~$CTER$_{\mathrm{Inst}}$) denotes the ratio of samples where the novel compounds are translated incorrectly, and \textit{aggregate-level} CTER (i.e.,$~$CTER$_{\mathrm{Aggr}}$) denotes the ratio of samples where the novel compounds suffer at least one incorrect translation when aggregating all 5 contexts. To calculate CTER, \cite{li2021compositional} manually construct a dictionary for all the atoms based on the training set, since each atom contains different translations. We also report character-level BLEU scores~\cite{DBLP:conf/acl/PapineniRWZ02} using SacreBLEU~\cite{DBLP:conf/wmt/Post18} as a supplement. For CFQ, we use exact match accuracy to evaluate model performance, where natural language utterances are mapped to meaning representations.

\subsection{Model Settings}
\label{sec:mst}

\vpara{Machine Translation.} We compare our method with competitive systems: (1) Transformer~\cite{vaswani2017attention}; (2) Transformer-Rela: only replaces absolute positional embedding with a relative one; (3) Transformer-Small: only decreases the number of encoder layers and decoder layers to 4, 4 respectively; (4) Transformer-Deep: only increases the number of encoder layers and decoder layers to 8, 8 respectively; (5) Bow~\cite{raunak2019compositionality}: uses bag-of-words pre-training; (6) SeqMix~\cite{guo2020sequence}: synthesizes examples to encourage compositional behavior; (7) Dangle~\cite{zheng2021disentangled}: adaptively re-encodes (at each time step) the source input.\footnote{We use the same variant reported by~\cite{zheng2021disentangled} (i.e., Dangle-EncDec (abs)) with absolute positional embedding.} (8) Proto-Transformer~\cite{yin2022categorizing}: integrates prototypes of token representations over the training set into the source encoding; (9) Transformer+CReg~\cite{yin2023consistency}: promotes representation consistency across samples and prediction consistency for a single sample; (10) DLCL~\cite{wang2019learning}: proposes an approach based on dynamic linear combination of layers (DLCL), and is one of the very popular EnocderFusion work. Our method is built on top of (1)-(3), i.e., LRF, LRF-Rela and LRF-Small.

\vpara{Semantic Parsing.} We compare our method with competitive systems: (1) LSTM+attention: introduces attention mechanism in LSTM~\cite{DBLP:journals/neco/HochreiterS97}; (2) Transformer~\cite{vaswani2017attention}; (3) Universal Transformer~\cite{DBLP:conf/iclr/DehghaniGVUK19}: combines the parallelizability and global receptive field of feed-forward sequence models; (4) Evolved Transformer~\cite{DBLP:conf/icml/SoLL19}: uses wide depth-wise separable convolutions in the early layers of both the encoder and decoder; (5) CGPS~\cite{li2019compositional}: leverages prior knowledge of compositionality with two representations, and adds entropy regularization to the encodings; (6) NSEN~\cite{DBLP:conf/nips/FreivaldsOS19}: is derived from the Shuffle-Exchange network; (7) T5-11B~\cite{DBLP:journals/jmlr/RaffelSRLNMZLL20}: treats every natural language processing task as a text-to-text problem, and is therefore suitable for the semantic parsing tasks. T5-11B is a T5 model with 11B parameters finetuned on CFQ; (8) T5-11B-mod~\cite{furrer2020compositional}: uses masked language model (MLM) pre-training together with an intermediate representation; (9) RoBERTa~\cite{liu2019roberta}: makes use of the RoBERTa-base model as the encoder and the randomly initialized Transformer decoder trained from scratch, where we use the same experimental settings of~\cite{zheng2021disentangled}; (10) HPD~\cite{NEURIPS2020_4d7e0d72}: proposes a novel hierarchical partially ordered set (poset) decoding paradigm; (11) Dangle~\cite{zheng2021disentangled}; (12) RoBERTa+CReg~\cite{yin2023consistency}; (13) LRF: builds on (9) with our method.

\begin{table}[!t]
    \centering
    \small
    \resizebox{0.9\linewidth}{!}{
    \begin{tabular}{@{}cc@{}c@{}c@{}c@{}}
      \toprule
    {\bf Model}  & \multicolumn{1}{c@{}}{\bf MCD1} & \multicolumn{1}{c@{}}{\bf MCD2} & \multicolumn{1}{c@{}}{\bf MCD3}& \multicolumn{1}{c@{}}{\bf MCD-Mean} \\
      \midrule
        LSTM+attention & 28.9 & 5.0 & 10.8 & 14.9 \\
        Transformer & 34.9 & 8.2 & 10.6 & 17.9 \\
        Universal Transformer & 37.4 & 8.1 & 11.3 & 18.9 \\
        Evolved Transformer & 42.4 & 9.3 & 10.8 & 20.8 \\
        CGPS & 13.2 & 1.6 & 6.6 & 7.1 \\
        NSEN & 5.1 & 0.9 & 2.3 & 2.8 \\
        T5-11B & 61.4 & 30.1 & 31.2 & 40.9 \\
        T5-11B-mod & 61.6 & 31.3 & 33.3 & 42.1 \\
        RoBERTa & 60.6 & 33.6 & 36.0 & 43.4 \\
        HPD & 72.0 & \bf 66.1 & \bf 63.9 & \bf 67.3 \\
        Dangle & \bf 78.3 & 59.5 & 60.4 & 66.1 \\
        RoBERTa+CReg & 74.8 & 53.3 & 58.3 & 62.1 \\
        \midrule
        LRF & 68.9 & 43.4 & 44.7 & 52.4 \\
       \bottomrule
    \end{tabular} 
    }
    \caption{Exact-match accuracy on different MCD splits of CFQ. Results are averaged over 3 random runs.}
\label{table:t3}
\end{table}

\subsection{Results on CoGnition}
\label{sec:roc}

The main results on CoGnition are shown in Table~\ref{table:t2}. We observe that: \textbf{(1)} LRF gives instance-level and aggregate-level CTERs of \textbf{20.0\%} and \textbf{50.3\%} respectively, with a significant improvement of \textbf{8.4\%} and \textbf{12.6\%} accordingly compared to the Transformer. Moreover, LRF outperforms most baseline models under the almost same parameter settings significantly, indicating fusing the syntactic and semantic information of sequences stored in different layers effectively is more beneficial to CG. Although Transformer+CReg achieves slightly better performance and contains fewer parameters, it is more complex and costly compared with LRF; \textbf{(2)} LRF, LRF-Rela and LRF-Small can deliver various performance improvements, demonstrating the general effectiveness of our method; \textbf{(3)} Since LRF brings some extra parameters (approximately 12M), we investigate to what extent the performance improvement is derived from the increase of model parameters. Transformer-Deep performs slightly better than Transformer on the CG-test set, indicating that only increasing model parameters is slightly useful but far from sufficient. In addition, LRF-Small contains fewer parameters and better performance than Transformer. 
Compared to SeqMix, the improvement of LRF is more significant (2.3\% vs 12.6\% aggregate-level CTER). SeqMix utilizes linear interpolation in the input embedding space to reduce representation sparsity, and we suppose that the  samples synthesized randomly may be unreasonable and harmful to model training; \textbf{(4)} It can be seen that Transformer is even slightly better than DLCL, indicating the inappropriate composition instead brings noise to significantly affect the model’s CG performance. Instead, we use the attention mechanism to fuse the most relevant parts of previous layers’ information at each layer in a flexible manner.

\subsection{Results on CFQ}
\label{sec:rocfq}

The main results on CFQ are presented in Table~\ref{table:t3}. We observe that: \textbf{(1)} RoBERTa is comparable to T5-11B,  T5-11B-mod and outperforms other baseline systems without pre-training except HPD, indicating that pre-trained language models are key to achieving good performance on CFQ, which is consistent with the conclusions in~\cite{furrer2020compositional}; \textbf{(2)} LRF substantially boosts the performance of RoBERTa (\textbf{43.4 $\rightarrow$ 52.4}), about \textbf{21\%} relative improvements, and is in fact superior to T5-11B and T5-11B-mod. It also outperforms other baseline systems without pre-training except HPD. This result demonstrates that pre-training as a solution to CG also has limitations, and also indicates that LRF is complementary to pre-trained models; \textbf{(3)} HPD performs better than Dangle, RoBERTa+CReg and LRF, achieving 67.3 exact match accuracy, which is highly optimized for the CFQ dataset. On the contrary, LRF, RoBERTa+CReg and Dangle are generally applicable to any seq2seq models for solving any seq2seq tasks including MT, as stated before. However, compared with the impressive performance on CoGnition, the improvements brought by LRF is relatively moderate, and even worse than Dangle. The underlying reason is connected a recent finding that compositionality in natural language is much more complex than the rigid, arithmetic-like operations~\cite{raunak2019compositionality,li2021compositional,zheng2021disentangled,DBLP:conf/acl/DankersBH22}. 
\begin{figure*}[!t]
      \centering
      \small
      \resizebox{0.8\linewidth}{!}{
      \subfloat[][The attention probabilities for 6-layer encoder.\label{fig4.1}]
      {\includegraphics[width=0.6\linewidth]{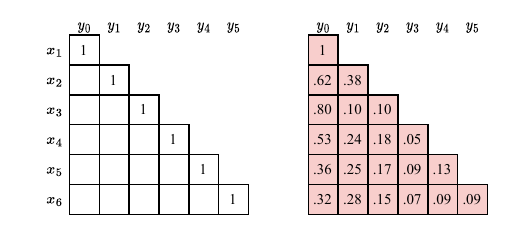}}
      \subfloat[][The attention probabilities for 6-layer decoder.\label{fig4.2}]
      {\includegraphics[width=0.6\linewidth]{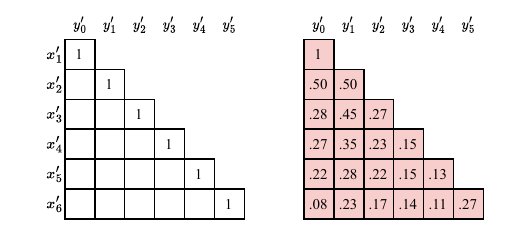}}
      }
      \caption{The differences between LRF and the Transformer in the process of fusing previous layers' information. $\{y_{0}, ..., y_{l} \}$ and $\{y_{0}', ..., y_{l}' \}$ are the output of the encoder and decoder layers $0 \sim l$ respectively. $\{x_{1}, ..., x_{l} \}$ and $\{x_{1}', ..., x_{l}' \}$ are the input of the encoder and decoder layers $1 \sim l$ respectively. $y_{0}$, $y_{0}'$ and $y_{<i}$, $y_{<i}'$ denote the input embedding and previous layers' outputs in the $i$-th encoder and decoder layer respectively. Red denotes the attention probabilities are learned by LRF.}
      \label{fig:fig4}
    \end{figure*}
MT is paradigmatically close to the tasks typically considered for testing compositionality in natural language, while our approach is more suitable for dealing with such scenarios.

\begin{table}[!t]
    \centering
    \resizebox{1.0\linewidth}{!}{
    \begin{tabular}{lll}
        \toprule
        \textbf{Model} & \textbf{$~$CTER$_{\mathrm{Inst}}\downarrow$} & \textbf{$~$CTER$_{\mathrm{Aggr}}\downarrow$} \\
        \midrule
        Transformer & 28.4\% & 62.9\% \\
        \quad+ Source Fuse-attention & 21.2\% (-7.2\%) & 53.3\% (-9.6\%) \\
        \quad+ Target Fuse-attention & 25.8\% (-2.6\%) & 58.8\% (-4.1\%)\\
        \quad+ Source \& Target Fuse-attention & \textbf{20.0\% (-8.4\%)} & \textbf{50.3\% (-12.6\%)} \\
        \bottomrule
    \end{tabular}}
    \caption{CTERs (\%) against the fuse-attention modules at different sides on the CG-test set.}
    \label{table:t4}
\end{table}

\section{Analysis}
\label{sec:ana}

In this section, we conduct in-depth analyses of LRF to provide a comprehensive understanding of the individual contributions of each component. For all experiments, we train a LRF (6-6 encoder and decoder layers) on the CoGnition dataset, unless otherwise specified.

\subsection{Do we need to introduce the fuse-attention module in both side?}
\label{sec:efmsts}
We argue that ``shallow'' residual connections with each layer of encoder or decoder lead to encoder or decoder RE problem respectively. Therefore, we are curious about whether LRF can alleviate both the encoder and decoder RE problems, and which one is more severe or affects the model’s ability of CG to a greater extent that is ignored in previous work. In this experiment, we investigate its influence on CoGnition. As shown in Table~\ref{table:t4}, we observe certain improvements (\textbf{-7.2\%} and \textbf{-2.6\%}$~$CTER$_{\mathrm{Inst}}$, \textbf{-9.6\%} and \textbf{-4.1\%}$~$CTER$_{\mathrm{Aggr}}$) when separately applying the fuse-attention modules at the encoder or decoder side. It suggests that our proposed method can alleviate the encoder and decoder RE problem respectively, and fusing information from previous layers back into the encoding or decoding process at each layer effectively can improve CG performance. 
Furthermore, their combination brings further improvement (\textbf{-8.4\%}$~$CTER$_{\mathrm{Inst}}$, \textbf{-12.6\%}$~$CTER$_{\mathrm{Aggr}}$), which illustrates that LRF can alleviate both the encoder and decoder RE problems and have cumulative gains. In addition, it is clear that the improvement from introducing the fuse-attention module at the encoder is more significant than that at the decoder, indicating that the encoder RE problem is more severe. The underlying reason is related to a recent finding that
the encoder has a greater impact on performance than the decoder~\cite{DBLP:conf/iclr/Kasai0PCS21,DBLP:conf/naacl/XuGLX21}. 

\begin{table}[!t]
    \centering
    \resizebox{0.9\linewidth}{!}{
    \begin{tabular}{cll}
        \toprule
        \textbf{Model} & \textbf{$~$CTER$_{\mathrm{Inst}}\downarrow$} & \textbf{$~$CTER$_{\mathrm{Aggr}}\downarrow$} \\
        \midrule
        Transformer & 28.4\% & 62.9\% \\
        Transformer-accu & 37.3\% (+8.9\%) & 71.4\% (+8.5\%) \\
        LRF-onlytop & 22.1\% (-6.3\%) & 52.7\% (-10.2\%) \\
        LRF & \textbf{20.0\% (-8.4\%)} & \textbf{50.3\% (-12.6\%)} \\
        \bottomrule
    \end{tabular}
    }
    \caption{CTERs (\%) on the CG-test set.}
    \label{table:t5}
\end{table}

\subsection{Do we need to fuse previous layers' information at each layer effectively?}

\begin{table*}
\centering
\small
\resizebox{0.95\linewidth}{!}{
\begin{tabular}{c|c|c}
\toprule
\textbf{Source} & \textbf{Transformer} & \textbf{LRF} \\ \midrule
\begin{tabular}[c]{@{}c@{}}\textbf{The waiter he liked}\\ wore each other's clothes.\end{tabular} & \begin{tabular}[c]{@{}c@{}}\includegraphics[height=3.2mm]{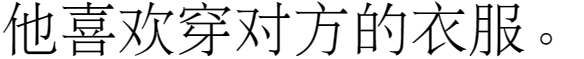}\\ (\textbf{He liked} to wear each other's clothes.)\end{tabular} & \begin{tabular}[c]{@{}c@{}}\includegraphics[height=3.2mm]{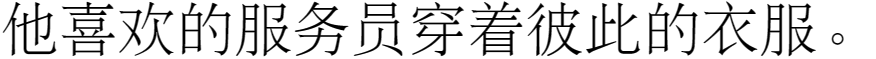}\\ (\textbf{The waiter he liked} wore each other's clothes.)\end{tabular} \\ \midrule
\begin{tabular}[c]{@{}c@{}}\textbf{The waiter he liked} came\\ by and chased the bully off.\end{tabular} & \begin{tabular}[c]{@{}c@{}}\includegraphics[height=3.2mm]{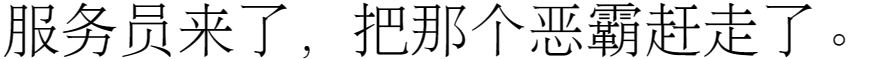}\\ (\textbf{The waiter} came by and chased the bully off.)\end{tabular} & \begin{tabular}[c]{@{}c@{}}\includegraphics[height=3.2mm]{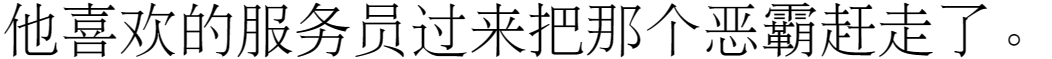}\\ (\textbf{The waiter he liked} came by and chased the bully off.)\end{tabular} \\ \midrule
\begin{tabular}[c]{@{}c@{}}\textbf{The waiter he liked} had a goal\\ to post a video every day.\end{tabular} & \begin{tabular}[c]{@{}c@{}}\includegraphics[height=3.2mm]{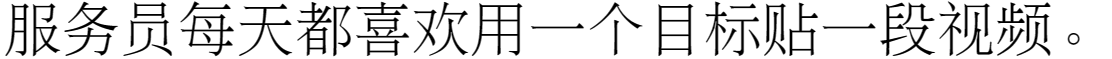}\\ (\textbf{The waiter liked} to have a goal to post a video every day.)\end{tabular} & \begin{tabular}[c]{@{}c@{}}\includegraphics[height=3.2mm]{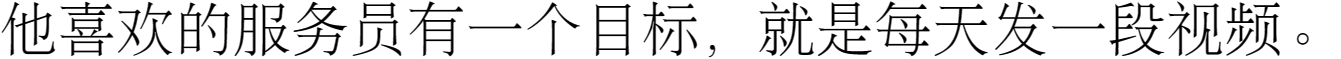}\\ (\textbf{The waiter he liked} had a goal to post a video every day.)\end{tabular} \\ \bottomrule
\end{tabular}}

\caption{
Example translations of Transformer vs. LRF.
The bold characters denote the novel compounds and corresponding translations. The English in parentheses represents the meaning of the model-generated Chinese.
}
\label{table:t7}
\end{table*}

We argue that it is important to introduce the fuse-attention module in \textbf{each layer}, since we view the emergence of RE problems as the process of gradually entangled representation from bottom to top of the Transformer layers. 
To validate this argument, we only introduce the fuse-attention module in the topmost layer of encoder and decoder (called LRF-onlytop). 
As shown in Table~\ref{table:t5}, LRF substantially boosts the performance of LRF-onlytop (\textbf{-2.1\%}$~$CTER$_{\mathrm{Inst}}$, \textbf{-2.4\%}$~$CTER$_{\mathrm{Aggr}}$). 
In addition, LRF-Small contains fewer parameters and better performance than LRF-onlytop.
As mentioned above, we use the fuse-attention module to achieve the goal of fusing previous layers' information at each layer \textbf{effectively}. To validate this argument, we also conduct a toy experiment on CoGnition. Specifically, each layer of the encoder and decoder accumulates corresponding previous layers' information (called Transformer-accu),\footnote{The input of encoder and decoder layer $i$ is $x_{i} = y_{0} + \cdots + y_{i-1}, x_{i}' = y_{0}' + \cdots + y_{i-1}'$, where $0 < i < L$ (see Figure~\ref{fig:fig4}).} rather than 
learn to fuse it adaptively like we do. Results are listed in Table~\ref{table:t5}. Transformer-accu even obtains worse generalization results than Transformer. It suggests that the ``simple'' combinations will instead bring noise to affect the model's CG performance.

To further demonstrate the fuse-attention module we introduced can fuse previous layers' information \textbf{effectively} and understand the individual contributions of previous layers' information for each layer, we visualize the attention probabilities of the fuse-attention modules in each encoder and decoder layer. Specifically, we train LRF on CoGnition and test on 680 (a batch) randomly selected examples of CG-test set, and then extract the attention probabilities of the fuse-attention modules. Ideally, the fuse-attention modules in different layer of the encoder and decoder should learn different and specific combinations of previous layers' information. 
In Figure~\ref{fig:fig4}, we observe that each layer of the encoder or decoder assigns different attention weights to information from previous layers as expected. This implies that the fuse-attention module in LRF can learn to fuse representations from previous layers at each layer effectively.

\subsection{Effects on Compositional Generalization}
\label{sec:ecg}

\begin{figure}[!t]
    \centering
    \includegraphics[width=0.9\linewidth]{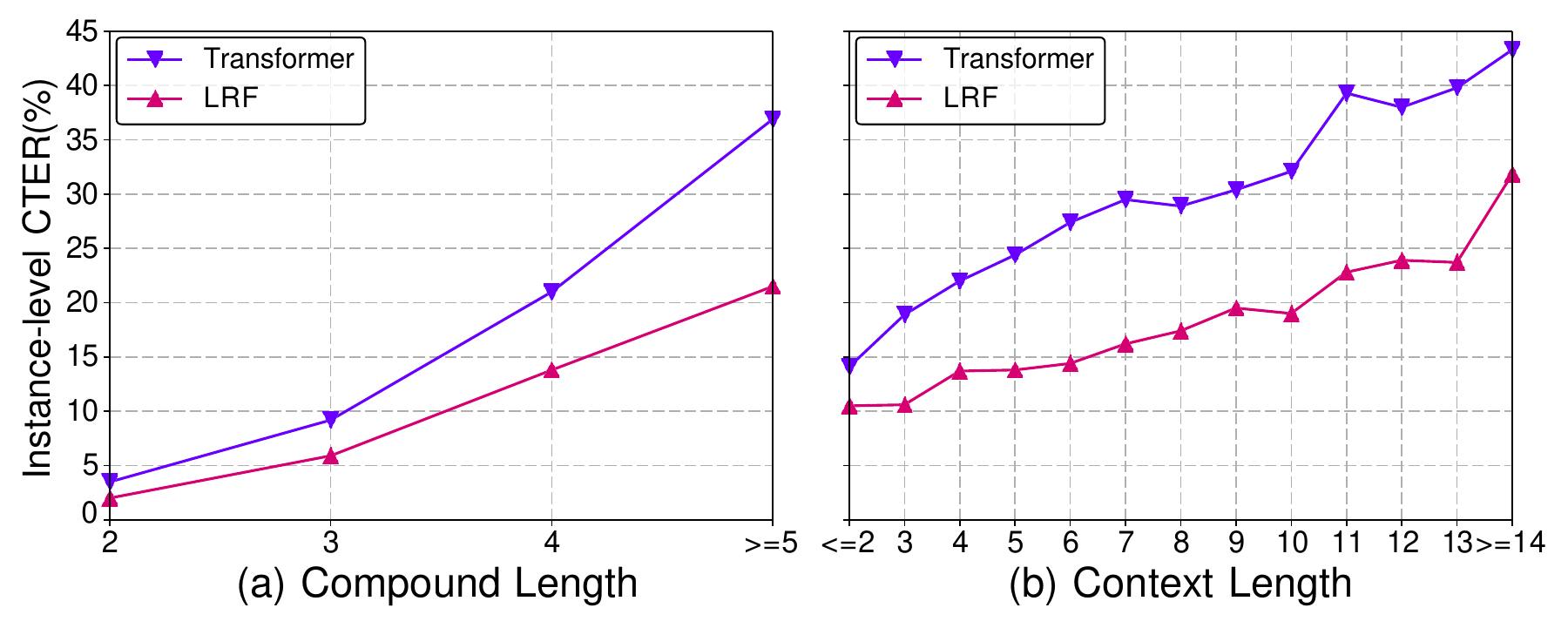}
    \caption{Instance-level CTER of LRF and Transformer over the different compound and context lengths.}
    \label{fig:fig5}
\end{figure}

\vpara{Compound Length and Context Length.} Longer compounds have more complex semantic information and longer contexts are harder to comprehend, making them more difficult to generalize~\cite{li2021compositional}. We classify the test samples by compound length and context length, and calculate the $~$CTER$_{\mathrm{Inst}}$. In Figure~\ref{fig:fig5}, we can observe that LRF generalizes better as the compound and context grows longer compared to Transformer. In particular, LRF gives a lower CTER by \textbf{11.5\%} over samples when the context leangth is longer than 13 tokens. It suggests that LRF can better captures the compositional structure of human language.

\begin{figure}[!t]
    \centering
    \includegraphics[width=0.68\linewidth]{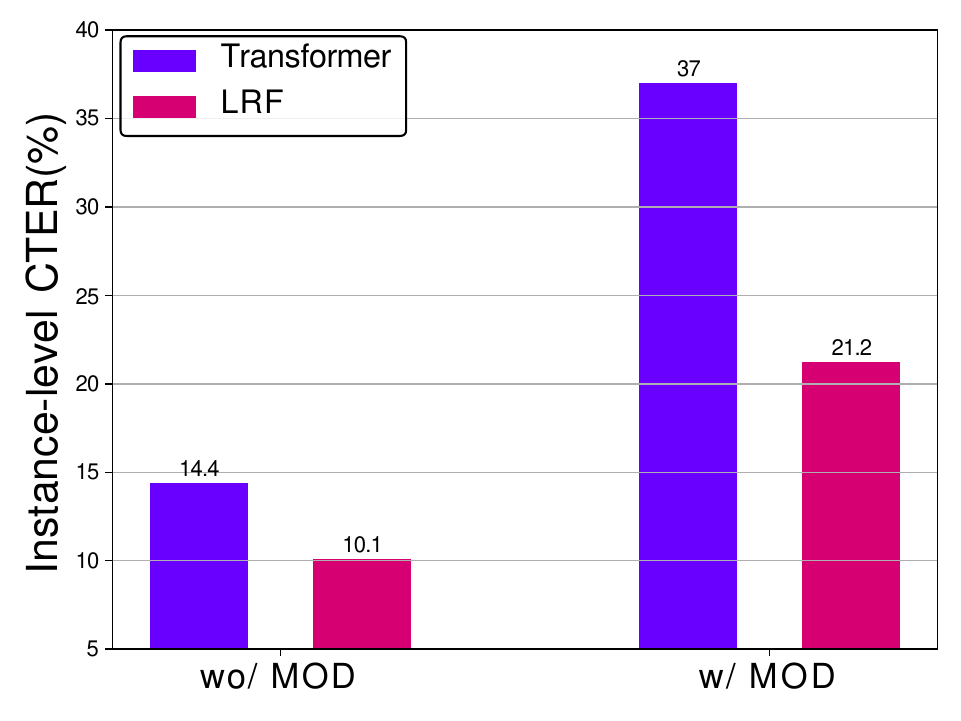}
    \caption{$~$CTER$_{\mathrm{Inst}}$ on compounds w/o and w/ MOD.}
    \label{fig:fig6}
\end{figure}

\vpara{Complex Modifier.} The postpositive modifier atom (MOD) is used to enrich the information of its preceding word (e.g., \emph{he liked} in the phrase \emph{lost the dog he liked}), which is challenging to translate due to word reordering from English to Chinese. We divide the test samples into two groups according to compounds with (w/) or without (wo/) MOD. In Figure~\ref{fig:fig6}, we observe that the advantage of LRF grows larger in translating the compounds with MOD, demonstrating its superiority in processing complex semantic composition. 

\vpara{Case Study.} We present 3 source examples containing a novel compound with MOD and 4 atoms, and their translations in Table~\ref{table:t7}. For all samples, correct translations denote that the novel compounds are translated correctly. LRF correctly translates the novel compounds across different contexts for all samples, while Transformer suffers from omitting different atoms. For example, the translation of \emph{the waiter} is omitted in the first example, \emph{he liked} is omitted in the second example and \emph{he} is omitted in the third example. Our results not only contain the correct compound translations but also achieve better translation quality, while Transformer makes various errors on unseen compositions.

\section{Related Work}
\label{sec:rw}

\vpara{Compositional Generalization.} CG has long played a prominent role in language understanding, explaining why we understand novel compositions of previously observed elements. Therefore, after realizing existing neural models still struggle in  scenarios requiring CG~\cite{lake2018generalization,keysers2019measuring,li2021compositional}, there have been various studies attempt to improve the model’s ability of CG, including data augmentation~\cite{andreas-2020-good,DBLP:conf/iclr/AkyurekAA21,DBLP:conf/naacl/YangZY22,li2023learning}, modifications on model architecture~\cite{li2019compositional,russin2019compositional,NyeS0L20,liu2020compositional,DBLP:conf/acl/LiuALLCLWZZ21,ZhengL21,HerzigB20,chaabouni2021can,wang2021structured,DBLP:conf/iclr/MittalRRBL22,lin2023learning}, intermediate representations~\cite{furrer2020compositional,herizgunlock21}, meta-learning~\cite{Lake19meta,ConklinWST20}, explorations on pre-trained language models~\cite{furrer2020compositional,zhou2023leasttomost}, auxiliary objectives~\cite{JiangB21indu}, and enriching semantic information at token-level~\cite{thrush2020compositional,akyurek2021lexicon,zheng2021disentangled,DBLP:conf/emnlp/YaoK22}. One line of research exploring how to alleviate the RE problems has attracted much attention. Our work is in line with it, we examine CG from a new perspective to solve it.

\vpara{Neural Machine Translation.} Recently, CG and robustness of Neural Machine Translation (NMT) have gained much attention from the research community~\cite{cheng2020advaug,xu2021addressing,lake2018generalization,li2021compositional}. \cite{raunak2019compositionality} propose bag-of-words pre-training for the encoder. \cite{guo2020sequence} propose sequence-level mixup to create synthetic samples. Recently, \cite{li2021compositional} introduce a practical benchmark dataset for analyzing fine-grained and systematic compositional ability of NMT, called CoGnition. \cite{DBLP:conf/acl/DankersBH22} argue that MT is a suitable and relevant testing ground to test CG in natural language. Based on this, \cite{zheng2021disentangled} propose to adaptively re-encode the source input at each time step.  Different from them, our method introduce a fuse-attention module in each encoder and decoder layer to fuse previous layers' information back into the encoding and decoding process effectively, which is inspired by previous explorations about analyzing how Transformer encoder and decoder perform~\cite{peters-etal-2018-deep,he2019hard,voita2019bottom,belinkov2020linguistic}.

\section{Conclusion}
\label{sec:ccl}

In this paper, we propose LRF, which learns to fuse previous layers' information back into the encoding and decoding process effectively through introducing a fuse-attention module at each encoder and decoder layer. Experiments on CoGnition and CFQ have shown the effectiveness of our proposal without any dataset or task-specific modification. To our knowledge, we are the first to explain why the RE problems exist and investigate how to fuse previous layers' information at each layer effectively to alleviate it, achieving better generalization results. We hope the work and perspective presented in this paper can inspire future related work.

\section{Acknowledgments}
We thank all the anonymous reviewers for their insightful and valuable comments. This work is supported by National key R\&D Program of China (Grant no.2022ZD0116101), the Key Support Project of NSFC-Liaoning Joint Foundation (Grant no. U1908216), and the Project of Research and Development for Neural Machine Translation Models between Cantonese and Mandarin (No. WT135-76).

\bibliography{aaai24}

\end{document}